\documentclass{article}
\usepackage{spconf,amsmath,graphicx,hyperref, amsfonts, algorithm, algpseudocode}
\newcommand{\myquad}[1][1]{\hspace*{#1em}\ignorespaces}
\usepackage{booktabs}   % For professional-looking tables (\toprule, \midrule, \bottomrule)
\usepackage{siunitx}    % For aligning numbers on the decimal point (S column type)
\usepackage{multirow}   % For cells that span multiple rows
\usepackage{graphicx}   % Required for \rotatebox if you need to rotate headers
\usepackage{lipsum}     % To generate dummy text to show the table placement
\usepackage{setspace}   % To set the line spacing in algorithm

\title{Can abstract concepts from LLM improve SLM performance?}
\name{Siddharth Tandon\thanks{At the time of writing the paper author was working as Strategic Cloud Engineer - AI in Google, Global Services Delivery team}}
\address{Google AI, Global Services Delivery}
\begin{document}
%\ninept
%
\maketitle
\begin{abstract}
Large language models (LLMs) excel at diverse tasks, but their deployment on resource-constrained devices remains challenging. Existing methods like quantization, pruning, and distillation can reduce memory footprint but often demand extensive experimentation and careful infrastructure design. Leveraging existing techniques for extracting high-level concepts (represented as steering vectors) from larger models, we investigate their transferability to smaller language models (SLM) during inference. We demonstrate through extensive experimentation that these concepts can be effectively transferred to smaller models, irrespective of their family (e.g., Phi, Llama, Qwen), leading to performance improvements across a wide range of tasks. Furthermore, we introduce inference-time scaling to enhance performance by dynamically adjusting the steering intensity which has resulted in a 7-15\% of accuracy improvement for Qwen3-0.6B.

\end{abstract}
%
% \begin{keywords}
% One, two, three, four, five
% \end{keywords}
%
\section{Introduction}
\label{sec:intro}

Large language models \cite{brown2020language} have recently reshaped the field of NLP with their remarkable performance across a range of complex language benchmarks \cite{bommarito2022gpt,wei2022emergent,bubeck2023sparks}, but there deployment on edge devices remains a challenge in terms of compute resources available. Hence there is constant research in the direction of building small linear models (e.g. Phi, llama, Gemma, etc..) to improve the performance using their large variants \cite{abdin2024phi, dubey2024llama, team2025gemma}. Improving the performance of small language models has been tackled through various mechanisms like pruning, distillation, quantization, but they require extensive experimentation and careful infrastructure design. 

Pruning is typically divided into three types: unstructured \cite{frankle2018lottery, sun2023simple, tanaka2020pruning} semi-structured \cite{frantar2023sparsegpt,meng2020pruning,ma2020image}, and structured \cite{wang2022trainability, ma2023llm}. Only structured pruning does not require special hardware or software whereas both unstructured and semi-structured pruning need the support of special hardware or software. Recent works \cite{yang2025wanda++, sreenivas2024llm} have further pushed the boundary of pruning LLM architecture networks without degradation of performance. 

The fast pace of LLM quantization research \cite{chen2024efficientqat, zhang2024leanquant} has led to improvement in accuracy, efficiency, and adaptability in LLMs. Methods like quantization-aware training (QAT) \cite{jeon2024l4q, chen2024efficientqat} and post-training quantization (PTQ) \cite{ding2023cbq, liu2024spinquant, zhang2024leanquant} are being refined to minimize performance loss during the conversion of model parameters from floating-point to lower bit-width integer formats which has helped in reducing the model footprint. This has also led to efficient fine tuning \cite{dettmers2023qlora, hu2022lora} of large language models on a resource constrained hardware.

Knowledge distillation is also a compression technique where the knowledge is transferred from larger (teacher) model to small (student) model. Many of the latest small language models are now distilled from their larger counterparts. For example all gemma 3 models \cite{team2025gemma} are trained with knowledge distillation. Similarly Llama 3.2 1B and 3B uses distillation technique from their larger counterpart i.e. Llama 3.1 8B and 70B

Recently some of the works \cite{turner2023steering,panickssery2023steering, zou2023representation, belitsky2025kv} tried to uncover the hidden capabilities of the model by explicitly modifying the internal state of the model on concepts where eliciting some of the behaviors (like emotions, honesty, bias, fairness) cannot be achieved with straight forward prompt engineering methods. \cite{turner2023steering} has introduced activation engineering (ActAdd) where the difference in activations of contrastive prompts are added to internal model activations to get desired behavior. Similarly \cite{zou2023representation} also uses the same idea but instead of directly adding the difference they capture the variance between the difference of contrastive prompts and then either add/ project on the model activations. Recently \cite{belitsky2025kv} has proposed cache steering where they make a one-time modification directly to the key-value cache of a Transformer model.

These methods demonstrate effective control over a single model's behavior, they primarily focus on steering the same model from which the activations were derived. They do not explore the transference of these conceptual representations to different, smaller models, especially those from other architectural families. This leaves a critical question unanswered:\textbf{ Can the sophisticated concepts learned by large, powerful models be extracted and used to enhance smaller language models?}

The work is built upon the core method developed by \cite{zou2023representation} for extracting the steering vectors and controlling the output. In this work, we mainly analyzed the impact of steering vectors across the model architectures and tried to bridge the gap by answering the question how robust steering output is and if steering outputs can make the small language models better. We hypothesize that steering vectors extracted from a large source model can be thought of as a basis of large concept space and act as "conceptual booster" for a wide range of smaller target models at inference time, eliminating the need for costly retraining or fine-tuning. We systematically investigate the robustness and effectiveness of this transfer process.
Our main contributions are as follows:
\begin{enumerate}
    \item We propose cross-architecture framework for transferring conceptual knowledge from large source models to smaller target models. 
    \item We propose inference-time scaling to boost the performance of smaller models 
    \item We analyze the concepts learned by different large models and their similarity to one another.
    \item We conduct extensive evaluations across different model families, including Phi, Llama, and Qwen to understand the significance of extracted steering outputs on a diverse set of downstream tasks
    
\end{enumerate}

\section{Related work}
\label{sec: relatedwork}
\textbf{Fine tuning }
Fine tuning is the most common way of adapting a language model to a domain or task and recent progress in Parameter efficient fine tuning approaches helps in adapting large pre-trained language models to new tasks without having to retrain all of the model's parameters. This is in contrast to full fine-tuning, which can be computationally expensive and require significant memory.
Notable PEFT approaches include \cite{hu2022lora}, adapter tuning \cite{zhang2023adalora}, prefix tuning \cite{li2021prefix}, P tuning \cite{liu2024gpt} and others. Some of the prominent strategies as surveyed in  \cite{wang2025parameter} :

\begin{enumerate}

\item Adapter tuning \cite{houlsby2019parameter, lin2020exploring}  involves adding new, trainable modules or parameters to the model while keeping the original pre-trained weights frozen. This significantly reduces the computational and storage costs of fine-tuning. 

\item Reparameterization methods: Techniques like LoRA (Low-Rank Adaptation) \cite{hu2022lora} reparameterize weight updates using low-rank matrices. During inference, these smaller, trained matrices are merged back into the original weights, meaning there is no additional computational latency.  

\item Selective Fine tuning: Instead of adding new parameters, this technique fine-tunes only a small, selected subset of the original model's parameters. A parameter mask is used to determine which weights to update, which can be either structured or unstructured.

\item Quantization PEFT: This technique first quantizes the pre-trained model (reduces the precision of its weights, for example, to 4-bit) and then applies an efficient tuning method. A popular example is QLORA \cite{dettmers2023qlora}, which combines 4-bit quantization with LoRA.
\end{enumerate}

Irrespective of all the recent advancement in parameter efficient fine tuning approaches this still requires an accelerator, managing memory for very large models, setting up pipelines to train a section of parameters to get optimal performance

\textbf{Representation Engineering}

Representation Engineering \cite{zou2023representation} focuses on understanding and manipulating the high-level representations of concepts like honesty, fairness, and power-seeking within a neural network. The goal is to develop methods to monitor and control the behavior of large language models, thereby improving their safety and reliability. This is done by first passing a pair of contrastive prompts to model, computing the difference of internal activation of both the prompts, and then capturing the variance of these differences 

For each prompt $s$ in the pair, retrieve the hidden state values from all layers typically for the last token. This results in a collection of hidden states, denoted as $H$ structured as:
$$
[\{H(s_0), H(s_1)\}, \{H(s_2), H(s_3)\}, \dots]
$$
Then proceed by computing the difference between the hidden states within each pair. This difference is then normalized. Formally, for a pair $\{H(s_i), H(s_{i+1})\}$, the difference $D$ is:
$$
D(s_i, s_{i+1}) = \text{normalize}(H(s_i) - H(s_{i+1}))
$$

Post this, construct a PCA model using these normalized hidden states difference vectors, and take the first component $PC1$. Although in practice we can take additional components but first component can give decent performance

Finally during inference for each prompt $s \in S_{test}$  capture the hidden states for all layers $[H(s_0), H(s_1), \dots]$ typically at the last token sequentially and then modify internal representations like below:
$$h_{l}^{*} = h_{l} + \lambda * PC1$$

% This lambda plays a very crucial role in deciding how much internal state activations to be modified. Too large values will break the model and too small values will not have any effect

\textbf{KV Cache Steering} 
KV Cache is another work which leverages steering vectors similar to \cite{zou2023representation} and \cite{turner2023steering} but instead of controlling the output of layer output they modify the cached key and value vectors at a target position of the KV cache 
\begin{align*}
V^*_{l} &= V_{l} + c^{v}S^{v}_{l} \\
K^*_{l} &= K_{l} + c^{k}S^{k}_{l}
\end{align*}

where $K_{l}, V_{l} \in \mathbb{R}^{H \times D_{h}}$ are the original cached key and value vectors at layer $l$, and $S^{k}_{l}, S^{v}_{l} \in \mathbb{R}^{H \times D_{h}}$ are the steering vectors calculated by forward pass contrastive prompts and taking mean of differences, and $c^{k}, c^{v} \in \mathbb{R}$ are scalar coefficients controlling the steering strength.

% --- THE GOLD STANDARD TABLE ---
% We use the `table*` environment so it spans both columns of the page.
% LaTeX will automatically float this to the top of the next available page.
\begin{table*}[t!]
    \centering % Center the table on the page
    \normalsize    % Use a slightly smaller font size for better fitting. Avoid \tiny or \resizebox if possible.

    \caption{
        A comprehensive comparison of performance for various child models, grouped by their parent architecture.
        We report two metrics: Baseline, Best Lambda ($\lambda$*), across three standard benchmarks. The results are reported upto 2 decimal places. 
        All values are performance scores (e.g., accuracy). 
    }
    \label{tab:my_main_results} % A label for cross-referencing in the text, e.g., using \ref{tab:my_main_results}

    \begin{tabular}{
        l                               % Left-aligned column for Model
        *{6}{S[table-format=1.1]}      % 12 columns for numbers aligned on the decimal point
    }
    \toprule % Top line of the table

    % --- HEADER ROWS ---
    % A single header for the merged model column.
    \multirow{2}{*}{\textbf{Model}} & \multicolumn{2}{c}{\textbf{GSM8K}} & \multicolumn{2}{c}{\textbf{MATH}} & \multicolumn{2}{c}{\textbf{ARC-c}} \\
    \cmidrule(lr){2-3} \cmidrule(lr){4-5} \cmidrule(lr){6-7}  % Adjusted cmidrule to match new column count
     & {Baseline} & {$\lambda$}* & {Baseline} & {$\lambda$}* & {Baseline} & {$\lambda$}* \\
    \midrule % Line separating header from data

    % --- DATA ROWS ---
    % Parent models are now section headers spanning all columns.
    \multicolumn{7}{l}{\textbf{microsoft/phi-4}} \\
    \qquad gemma-2-2b-it & 51.09 & 49.73 & 6.82 & 6.68 & 70.30 & \textbf{70.64} \\
    \myquad[2] gemma-2-9b-it & 83.24 & 83.24 & 23.08 & \textbf{23.16} & 90.01 & \textbf{90.18} \\
    \myquad[2] Llama-3.1-8B-Instruct & 80.28 & \textbf{81.04} & 29.62 & 29.46 & 84.64 & \textbf{84.98} \\
    \myquad[2] Llama-3.2-1B-Instruct & 36.84 & \textbf{36.92} & 10.68 & \textbf{10.78} & 52.13 & 50.17 \\
    \myquad[2] Llama-3.2-3B-Instruct & 68.53 & \textbf{68.76} & 25.80 & 25.74 & 74.74 & \textbf{75.08} \\
    \myquad[2] Qwen2-7B-Instruct & 81.72 & 80.97 & 28.06 & \textbf{28.26} & 80.35 & \textbf{81.05} \\
    \myquad[2] Qwen3-0.6B & 48.97 & \textbf{49.27} & 5.40 & \textbf{5.42} & 21.24 & 20.22 \\
    \midrule

    \multicolumn{7}{l}{\textbf{Qwen/Qwen2.5-14B-Instruct}} \\
    \myquad[2] gemma-2-2b-it & 51.09 & 49.81 & 6.16 & \textbf{6.34} & 70.30 & \textbf{70.56} \\
    \myquad[2] gemma-2-9b-it & 83.24 & 83.16 & 19.28 & \textbf{19.62} & 90.01 & 89.67\\
    \myquad[2] Llama-3.1-8B-Instruct & 80.28 & \textbf{80.51} & 29.36 & \textbf{29.56} & 84.64 & \textbf{85.06} \\
    \myquad[2] Llama-3.2-1B-Instruct & 37.83 & 36.99 & 19.92 & 19.76 & 50.59 & 50.34 \\
    \myquad[2] Llama-3.2-3B-Instruct & 68.38 & \textbf{68.76} & 30.62 & \textbf{30.64} & 75.85 & 73.72 \\
    \myquad[2] Qwen2-7B-Instruct & 81.72 & \textbf{81.95} & 20.34 & \textbf{21.5}& 80.37 & \textbf{81.74} \\
    \myquad[2] Qwen3-0.6B & 48.97 & \textbf{50.64} & 5.32 & \textbf{5.54} & 21.24 & \textbf{24.40}\\
    \midrule

    \multicolumn{7}{l}{\textbf{Mistral-7B-Instruct-v0.3}} \\
    \myquad[2] gemma-2-2b-it & 51.09 & 50.87 & 7.44 & 7.44 & 70.30 & 70.13 \\
    \myquad[2] gemma-2-9b-it & 83.24 & \textbf{84.00} & 27.08 & \textbf{27.10} & 90.01 & 89.76 \\
    \myquad[2] Llama-3.1-8B-Instruct & 80.28 & 79.37 & 32.38 & 31.72 & 84.64 & 82.42  \\
    \myquad[2] Llama-3.2-1B-Instruct & 37.22 & 36.92 & 16.66 & \textbf{16.90} & 50.08 & \textbf{50.76} \\
    \myquad[2] Llama-3.2-3B-Instruct & 68.46 & 68.46 & 23.70 & \textbf{23.76} & 75.76 & 74.91 \\
    \myquad[2] Qwen2-7B-Instruct & 81.72 & 81.27 & 28.78 & \textbf{28.94} & 80.37 & \textbf{81.05} \\
    \myquad[2] Qwen3-0.6B & 48.97 & \textbf{49.88} & 6.08 & \textbf{11.10} & 21.24 & 19.88 \\
    
    \bottomrule % Bottom line of the table
    \end{tabular}
\end{table*}

\section{Approach}
\label{sec: approach}
Our approach focuses on transferring conceptual knowledge from a large language model to a more computationally efficient small target model without fine tuning the model

\subsection{Cross architecture framework pipeline}
\label{ssec:cross_arch}

We have used a pipeline leveraging decoder only transformer neural networks \cite{vaswani2017attention}. We use same logic as proposed by \cite{zou2023representation} to extract the representation vectors corresponding to specific concepts from the internal activation space of the source model. The intensity of this steering is controlled by a scalar hyperparameter, $\lambda$. To determine the optimal value for this coefficient, we conduct a search over a range of $\lambda$ values, evaluating the performance of the modified target model on a held-out validation set. The best-performing $\lambda$ is then fixed, and we report the final accuracy and other relevant metrics on a completely unseen test set to ensure an unbiased evaluation of our pipeline's effectiveness. Overall this is a three step pipeline to effectively transfer the concepts across model families
\begin{itemize}
    \setlength\itemsep{0em}
    \item Extract steering vectors using algorithm \ref{alg:steering}
    \item Identify $\lambda$ using algorithm \ref{alg:lambda}
    \item Evaluation on test set and report final metrics using Algorithm \ref{alg:evaluation}
\end{itemize}

Algorithm \ref{alg:steering} objective is to generate the steering vectors using the large language model, its corresponding tokenizer, and a small dataset. We have used the same code as provided by \cite{zou2023representation}. This gives us the steering vectors which will guide the small language model output in right direction

Algorithm \ref{alg:lambda} tries to identify the best $\lambda$ on validation dataset across the layers for various $\lambda$ values. Given the different width and depth of transformer architectures across model families, we employ a greedy layer-to-layer mapping ($n \rightarrow n$). If the source model has fewer layers than the target model, steering is applied only to the corresponding initial layers in the target model, and similarly when the source has more layers than the target. \ref{tab:my_main_results} show that even with such a greedy approach we are able to get increase in performance. Identify the right mapping between layers of different model is a future work

Once we have identified the best $\lambda$ we can run Algorithm \ref{alg:evaluation} to evaluate on test set. We have proposed to use inference time scaling with various $\lambda$ to increase the performance further. Once we have the results from various $\lambda$ we can perform the mode or any other metric depending on the dataset to report the final results
\begin{algorithm}[H]
\caption{Extract steering vectors}\label{alg:steering}
\begin{algorithmic}[1]
\Require Large language Model $M_l$ and corresponding tokenizer $T_l$, Dataset $D$, Utility function to extract steering vectors $f(Model, Tokenizer, Data)$, Data preparation function $P(Dataset, tokenizer, mode)$
\Ensure Steering Vectors $S \in \mathbb{R}^{L X D}$ \Comment{L: number of layers and D: model dimension}
\setstretch{1.2}
\State $D_{train} \leftarrow P(D, T_l, mode={train})$ 
\State $S \leftarrow f(M_l, T_l, D_{train})$
\State \Return $S$
\end{algorithmic}
\end{algorithm}

\begin{algorithm}[H]
\caption{Identify $\lambda$}\label{alg:lambda}
\begin{algorithmic}[1]
\Require Small language Model $M_s$ and corresponding tokenizer $T_s$, Dataset $D$, Data preparation function $P(Dataset,tokenizer, mode)$, Evaluation Metric function $E$, Steering vector $S$ from \ref{alg:steering}, Utility function to apply steering vectors to generated output of small language model $g(model, tokenizer, data, activations)$ 
\Ensure $\lambda_{\text{best}}$
\setstretch{1.2}
\State $D_{val} \leftarrow P(D, T_s, mode={'val'})$ 
\For{$\lambda \leftarrow \{0.01, \dots, 0.09\} \cup \{0.1, \dots, 1.0\}$}
    \State Initialize dictionary $act$
    \For{$layer \leftarrow 0$ to $|M_S|$ }  \Comment{$|M_S|$: Number of layers}
        \If{$layer \notin S$}
            \State $act[layer] \leftarrow [0]$
        \Else
            \State $act[layer] \leftarrow \lambda \cdot S[layer]$
        \EndIf
    \EndFor

    \State $outputs[\lambda] \leftarrow g(M_s, T_s, D_{val}, act)$
    
\EndFor
\State $\lambda_{\text{best}} \leftarrow \operatorname{argmax}_{\lambda} E(outputs[\lambda], D_{val}^{target})$

\State \Return $\lambda_{\text{best}}$
\end{algorithmic}
\end{algorithm}

\begin{algorithm}[t]
\caption{Evaluation on test set}\label{alg:evaluation}
\begin{algorithmic}[1]
\Require Small language Model $M_s$ and corresponding tokenizer $T_s$, Dataset $D$, Data preparation function $P(Dataset,tokenizer, mode)$, Evaluation Metric function $E$, Steering vector $S$ from \ref{alg:steering}, $\lambda_{\text{best}}$ from \ref{alg:lambda}, Utility function to apply steering vectors to generated output of small language model $g(model, tokenizer, data)$, $WITH\_ITS$: a boolean flag if inference time scaling is enabled 
\Ensure $results$
\setstretch{1.2}
\State $D_{test} \leftarrow P(D, T_s, mode={'test'})$ 
\If{$\mathit{WITH\_ITS}$}
    \State $\lambda_{\text{range}} \leftarrow \{0.01, \dots, 0.09\} \cup \{0.1, \dots, 1.0\}$
\Else
    \State $\lambda_{\text{range}} \leftarrow [\lambda_{\text{best}}]$
\EndIf

\For{$\lambda \leftarrow \lambda_{\text{range}}$}
    \State Initialize dictionary $act$
    \For{$layer \leftarrow 0$ to $|M_S|$ }  \Comment{$|M_S|$: Number of layers}
        \If{$layer \notin S$}
            \State $act[layer] \leftarrow [0]$
        \Else
            \State $act[layer] \leftarrow \lambda \cdot S[layer]$
        \EndIf
    \EndFor

    \State $outputs[\lambda] \leftarrow g(M_s, T_s, D_{test})$
    
\EndFor
\State $results \leftarrow \operatorname{mode}(outputs)$
    
\State \Return $results$

\end{algorithmic}
\end{algorithm}

\begin{figure*}[htb]
\centering

% Subfigure (a)
\begin{minipage}[b]{0.98\linewidth}
    \centering
    \includegraphics[width=\linewidth]{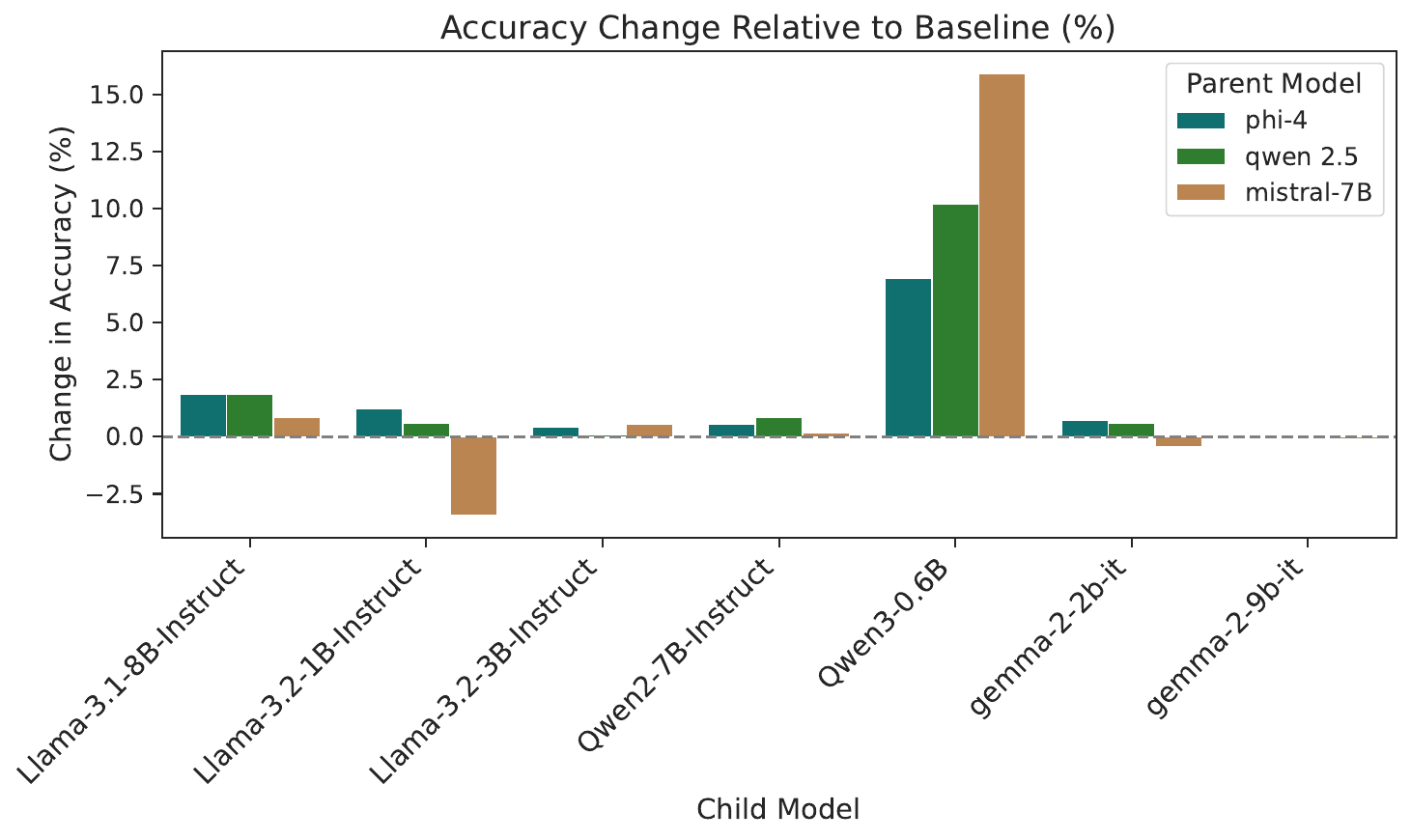}
    % \vspace{5pt} % Adjust space between image and text
    % \textbf{(a)}
\end{minipage}

\caption{Change in accuracy by Inference time scaling with baseline using GSM8k dataset. For each child model the three bars represent change in accuracy using different parent model.}
\label{fig:acc_its_change}
\end{figure*}

\section{Experiments}
\label{sec:experimnets}

\textbf{Datasets}: We have used following datasets: GSM8K \cite{cobbe2021gsm8k}, MATH \cite{hendrycksmath2021}, ARC-Challenge \cite{clark2018think}. MATH and GSM8k are mathematical datasets, whereas ARC-C provides grade-school level, multiple-choice science questions. Each dataset is used by both parent model for generating the steering output and controlling the output of child model. We have used the reasoning/ solution provided in MATH and GSM8k dataset  as it is whereas for ARC-C we have used Gemini model \cite{team2023gemini} to generate the reasoning. Training is done in a zero shot manner, whereas during inferencing we have have leveraged 5 few shots for each of the dataset for calculating both baseline results as well as steered results (\ref{tab:my_main_results}). For GSM8K we have used the few shots provided by \cite{NEURIPS2022_8bb0d291}, for MATH we have randomly picked 5 examples and excluded them during training, whereas for ARC-C we have created few shot examples using gemini \cite{team2023gemini}. Refer \ref{appendix:prompt} for details on prompts.

\textbf{Models}: We have evaluated the entire pipeline on various model architectures from hugging face of varying sizes. Parent model is from 3 model families(phi, Qwen, mistral) with 7B and 14B in size whereas child model size ranges from 0.6B to 9B of various model families (gemma, meta-llama, Qwen). Refer Appendix \ref{appendix:models}

As part of experiment we have used 2 separate a2-highgpu-1g machines on Google cloud with 1 NVIDIA Tesla A100 GPU on each machine. Each experiment first calculate the baseline values and then compute the steered output. The variation in the baseline values across child for each dataset is majorly because of experiments executing on different machine. We havent performed any hyper-parameter tuning for any of the experiments. For each dataset we have used the exact same prompt (refer Appendinx \ref{appendix:prompt}), few-shot, temperature, etc for all the parent child combinations.

\subsection{Can we improve the performance of smaller models?}
\label{ssec:exp_perf}
We evaluated the performance of several SLMs (ranging from 0.6B to 9B parameters) across three diverse reasoning datasets: GSM8K, MATH, and ARC-Challenge. For each child model, we compare the Baseline Accuracy (standard few-shot inference) against the Steered Accuracy using the optimal steering coefficient ($\lambda_{\text{best}}$) determined via Algorithm \ref{alg:lambda}. Refer appendix \ref{appendix:best_lambda} for $\lambda_{\text{best}}$ identified based on parent-child combination. As detailed in Table \ref{tab:my_main_results}, the application of concept steering consistently yields an increase in accuracy across all model architectures and datasets. 

The performance gains are observed regardless of the child model's family (e.g., Llama, Gemma, Qwen) or the parent model from which the concepts were extracted (e.g., $\text{Phi-4}$, $\text{Qwen2.5}$, $\text{Mistral}$). This confirms the architecture-agnostic nature of the transferred concepts

\subsection{Similarity of concepts}
\label{ssec:exp_sim}
\begin{figure*}[htb]
\label{fig:correlation}
\centering
% Subfigure (a)
\begin{minipage}[b]{0.85\linewidth}
    \centering
    \includegraphics[width=\linewidth]{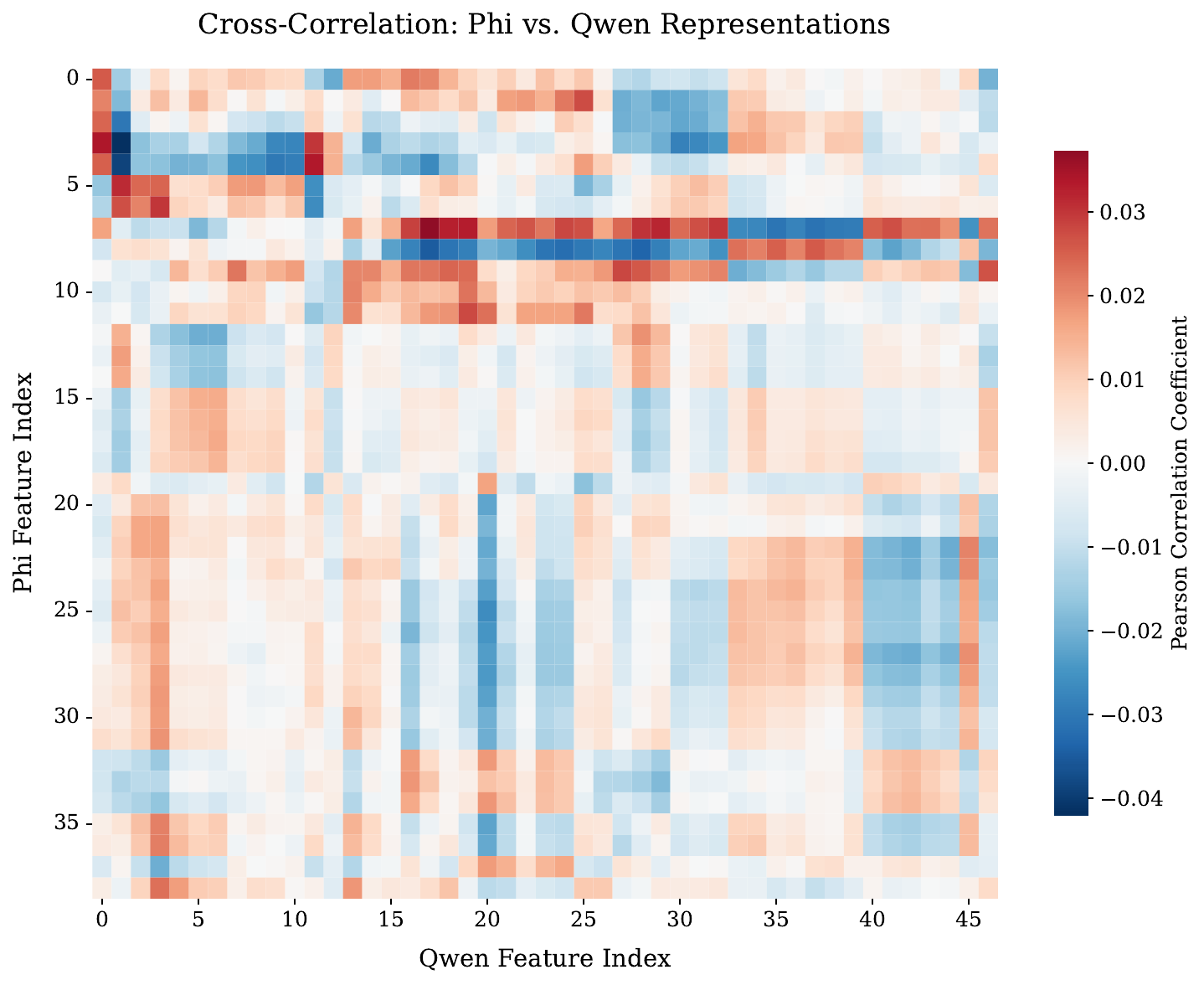}
    % \caption*{(a) Result 1}
    \vspace{5pt} % Adjust space between image and text
    \textbf{(a)}
\end{minipage}% <--- PERCENT SIGN PREVENTS NEW PARAGRAPH
\hfill
% Subfigure (b)

\caption{Correlation of variance captured by phi and qwen representation for each layer for arc-c dataset}
\label{fig:sim_comcepts}
\end{figure*}

To investigate the geometric alignment of the steering vectors across different parent models, we analyzed the linear relationship between the first PCA components of their layer-wise representations. Figure \ref{fig:sim_comcepts} illustrates the cross-correlation heatmap between the feature spaces of Phi-4 and Qwen 2.5 when processing the ARC-Challenge dataset.

We observed negligible correlation across all layers, suggesting that the primary direction of variance in the high-level concept space are architecture-specific. Future work could explore if a linear combination or a learned projection of these vectors yields a stronger alignment and further performance gains.

\subsection{Inference Time Scaling}
\label{ssec:exp_its}
Instead of using only the single fixed $\lambda_{\text{best}}$, Inference Time scaling involves running the small language model across a predefined range of steering coefficients ($\lambda_{range}$). As detailed in Algorithm \ref{alg:evaluation}, when the $WITH\_ITS$ flag is enabled, we iterate over $\lambda \in \{0.01, \dots, 0.09\} \cup \{0.1, \dots, 1.0\}$. For each input prompt in the test set, the target model generates a prediction. The final, aggregated result for that input is then determined by applying the mode of all the outputs generated across $\lambda_{range}$.

Figure \ref{fig:acc_its_change} compares the Accuracy Change Relative to Baseline (\%) for Inference Time Scaling across multiple child models and parent architectures. The figure \ref{fig:acc_its_change}, demonstrates that the result often provides a small, but significant, additional performance uplift beyond what is achieved with the optimal static steering coefficient. While positive scaling is observed across diverse families (Llama, Qwen, Gemma), the magnitude of improvement is particularly pronounced for the smallest model, Qwen3-0.6B, which sees gains of 7–15\% across various parent models. However, the benefits are not uniform; we observe isolated cases of performance decrease, such as with Llama-3.2-1B-Instruct steered by Mistral-7B, suggesting that parent-child compatibility remains a factor for specific architectures. for additional results refer Appending \ref{appendix:additional_its_results}

\begin{figure*}[htb]
\label{fig:lambda_results}
\centering
% Subfigure (a)
\begin{minipage}[b]{0.45\linewidth}
    \centering
    \includegraphics[width=\linewidth]{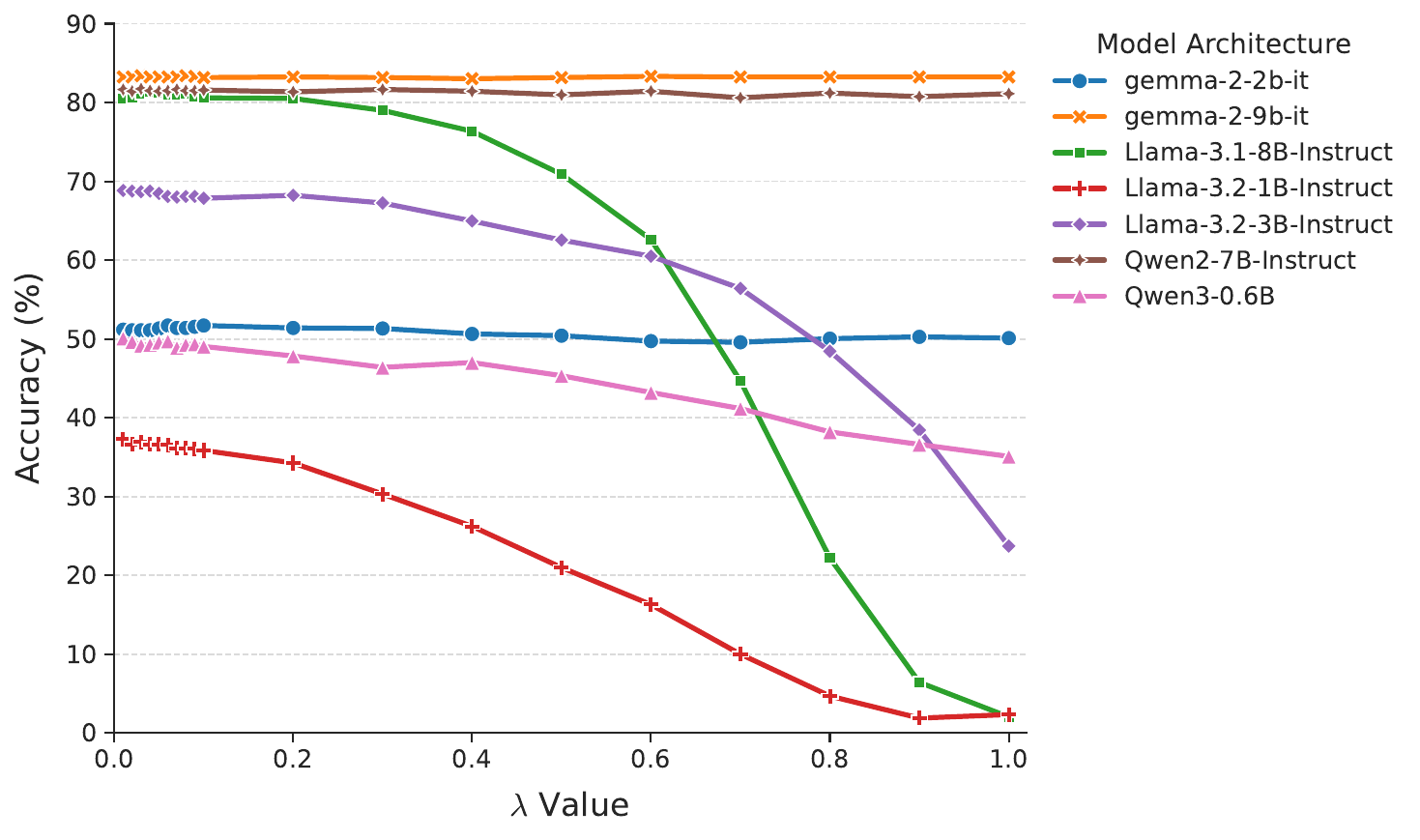}
    % \caption*{(a) Result 1}
    \vspace{5pt} % Adjust space between image and text
    \textbf{(a)}
\end{minipage}% <--- PERCENT SIGN PREVENTS NEW PARAGRAPH
\hfill
% Subfigure (b)
\begin{minipage}[b]{0.45\linewidth}
    \centering
    \includegraphics[width=\linewidth]{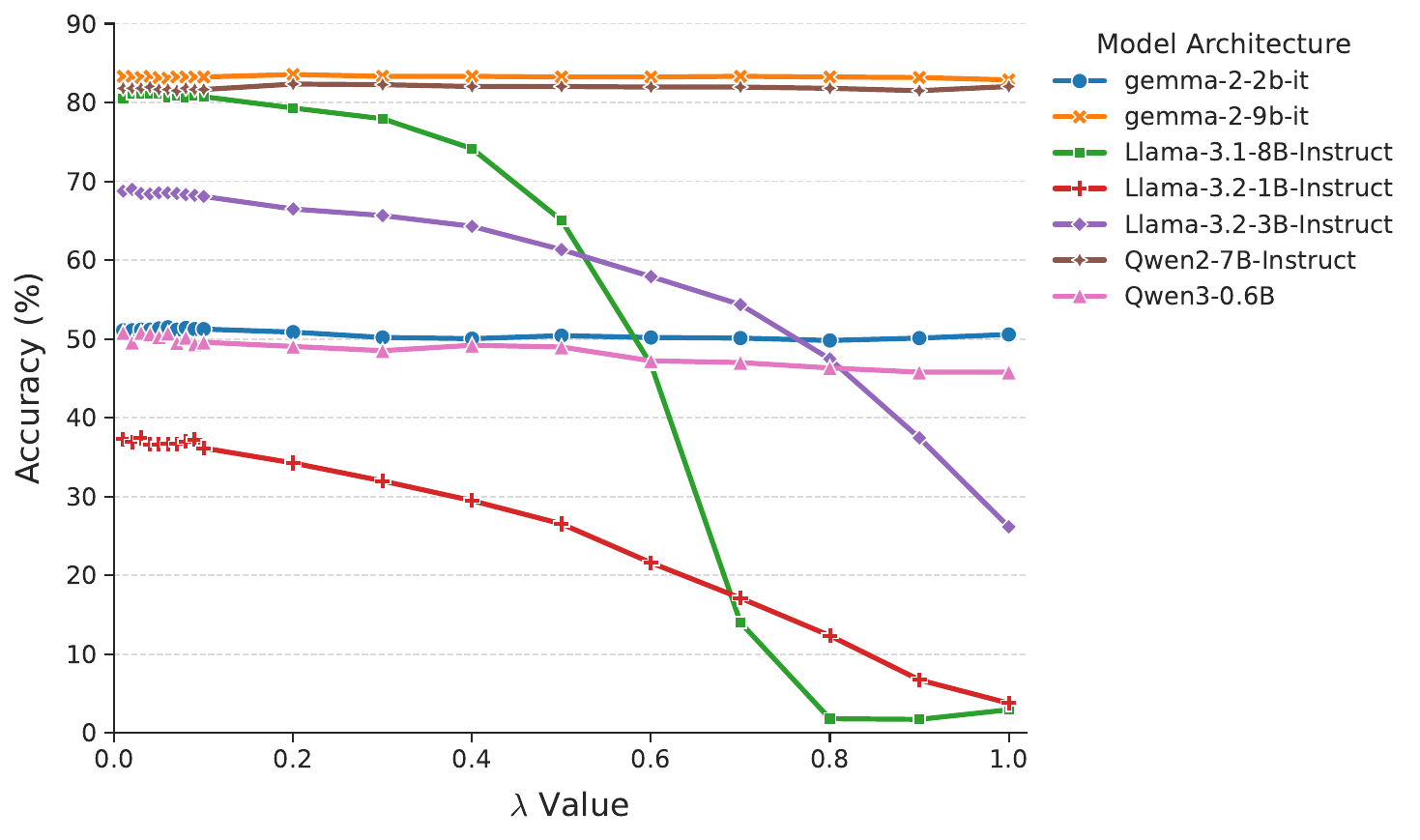}
    % \caption*{(b) Result 2}
    \vspace{5pt} % Adjust space between image and text
    \textbf{(b)}
\end{minipage}% <--- PERCENT SIGN PREVENTS NEW PARAGRAPH
\hfill
% Subfigure (c)
\begin{minipage}[b]{0.45\linewidth}
    \centering
    \includegraphics[width=\linewidth]{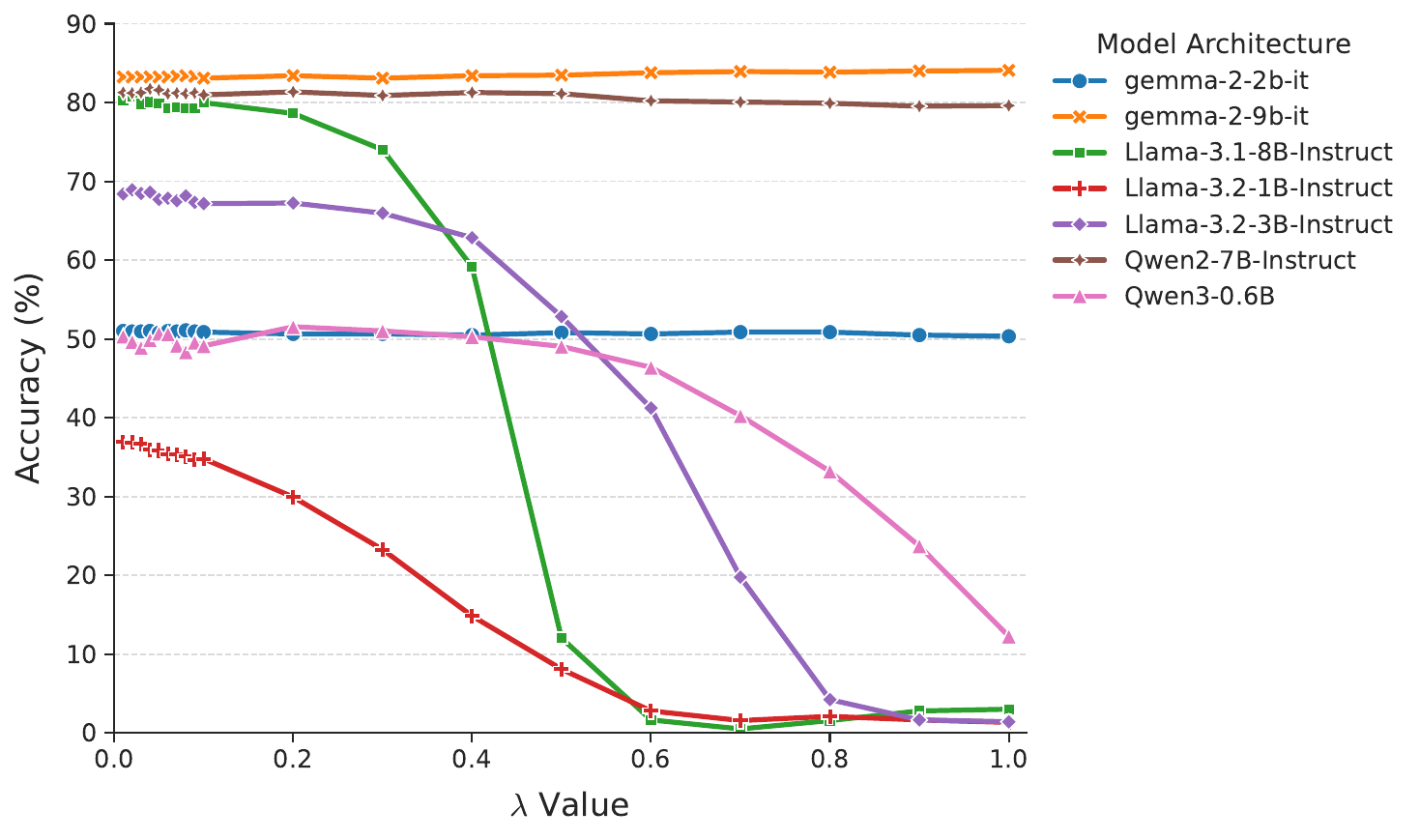}
    \vspace{5pt} % Adjust space between image and text
    \textbf{(c)}
\end{minipage}

\caption{Impact of $\lambda$ on various child model for GSM8k dataset when used with different parent models. (a) refers to  phi-4, (b) refers to Qwen2.5 14b-Instruct, (c) refers to Mistral-7B-Instruct-v0.3}
\label{fig:lambda_robust}
\end{figure*}

\subsection{Robustness of $\lambda$}
\label{ssec:exp_lambda}
We have experimented with various $\lambda \in \{0.01, \dots, 0.09\} \cup \{0.1, \dots, 1.0\}$ for all parent models and child models. Figure \ref{fig:lambda_robust} demonstrates the impact of $\lambda$ on various datasets across different parent-child combinations 

We observe a trend where performance is robust at lower values of $\lambda$ (typically $\lambda <= 0.3$) but tends to degrade as the steering intensity increases beyond this range. This suggests that while transferring concepts can boost performance, excessive modification of the activation space can disrupt the child model's internal representations, leading to a loss in accuracy.

Impact of Child Model:The robustness appears correlated with the child model architecture. As shown in Figure \ref{fig:lambda_robust}, Gemma models and Qwen2-7B maintain consistency across various $\lambda$ values. Llama models exhibit high sensitivity to $\lambda$ across all parent models. Notably, even the smaller Gemma-2-2B-IT demonstrates superior resiliency to performance degradation at large $\lambda$ values compared to the significantly larger Llama-3.1-8B, suggesting that the Gemma architecture is intrinsically more robust to external steering vectors. 

\section{Limitations}
\label{sec:limitaitons}
Our current framework employs a greedy approach of mapping layer from a large model to small model i.e. $n \rightarrow n$ which is is sub-optimal. The lack of a sophisticated, cross-architecture layer-mapping strategy represents a key limitation, as the optimal alignment of concepts across different network depths remains an open research question. Similarly determining one single $\lambda$ for entire model is also greedy in nature  and depends on validation dataset whereas in reality this should be a function of model depth and layer index. A more advanced method is needed to dynamically figure out $\lambda$ per layer and a dynamic layer-mapping strategy, which would allow for a more granular and effective transfer of conceptual knowledge tailored to each layer's specific contribution.

\section{Conclusion}
\label{sec:conclusion}
Our proposed cross-architecture framework effectively transfers conceptual knowledge—in the form of steering vectors—from a large source model to a smaller target model during inference. We empirically demonstrated the robustness and transferability of these concepts, showing that they act as a "conceptual booster" leading to performance improvements across various models (Phi, Llama, Qwen) and datasets (GSM8K, MATH, ARC-C). Furthermore, we introduced Inference-Time Scaling to dynamically optimize performance by adjusting the steering intensity $\lambda$. This work confirms our hypothesis that sophisticated concepts learned by large models can be leveraged to significantly enhance the capabilities of smaller, more efficient models without requiring costly fine-tuning. For future work, exploring a more sophisticated layer-mapping strategy between different model architectures and dynamically figuring out $\lambda$ per layer will be a key step in further optimizing this transfer process.

\vfill\pagebreak

\bibliographystyle{IEEEbib}
\bibliography{strings,refs}

% --- Start of Appendix ---
\clearpage
\onecolumn % Switch entire document to one column from here on
\appendix

\section{Experimental Prompts} \label{appendix:prompt}
\subsection{MATH Prompts}\label{appendix:mathprompts}

Table \ref{tab:math_prompt} shows the exact system prompt used for MATH dataset for all the models

\begin{table}[h!] % The * tells LaTeX to span both columns
    \centering
    \caption{System prompt used for MATH dataset during inferencing.}
    \label{tab:math_prompt}
    \renewcommand{\arraystretch}{1.5}
    % Use \linewidth or \textwidth to fill the newly expanded space
    \begin{tabular}{ | p{0.95\textwidth} | }
        \hline
        \textbf{System Prompt} \\
        \hline
        % The { brackets } scope the \obeylines command just to this cell
        {\obeylines
        You are an expert mathematician and a meticulous problem solver. Your task is to solve the following math problem from the MATH dataset. Follow these instructions carefully:
1. Understand the Problem: Read the problem statement carefully and identify the key information, the question being asked, and any constraints.
2. Formulate a Plan: Outline the steps you will take to solve the problem. State the mathematical concepts, formulas, or theorems you will use.
3. Show Your Work: Execute your plan step-by-step, showing all your reasoning and calculations clearly. This is crucial for understanding your thought process.
4. Final Answer: After your detailed solution, clearly state the final answer in a box. The format for the final answer should \boxed{answer}.
\par\vspace{\baselineskip}
Let's break down your response structure:
\par\vspace{\baselineskip}
- Step-by-step thinking: Present your reasoning in a logical and sequential manner. Explain each step of your calculation.
- Clarity and Precision: Use precise mathematical language and notation.
- Final Answer Encapsulation: The final numerical or symbolic answer must be enclosed in \boxed{answer}. 
} % End of \obeylines scope
\\
        \hline
    \end{tabular}
\end{table}

\clearpage
\subsection{GSM8k Prompts} \label{appendix:gsm8kprompts}

Table \ref{tab:gsm8k_prompt} shows the exact system prompt used for GSM8k dataset for all the models

\begin{table}[h!] % The * tells LaTeX to span both columns
    \centering
    \caption{System prompt used for GSM8k dataset during inferencing.}
    \label{tab:gsm8k_prompt}
    \renewcommand{\arraystretch}{1.5}
    % Use \linewidth or \textwidth to fill the newly expanded space
    \begin{tabular}{ | p{0.95\textwidth} | }
        \hline
        \textbf{System Prompt} \\
        \hline
        % The { brackets } scope the \obeylines command just to this cell
        {\obeylines
        You are an expert AI assistant specializing in solving grade-school math word problems (GSM8K). Your primary goal is to provide a clear, accurate, and step-by-step solution to the given problem. You must act as a meticulous and logical thinker.
\par\vspace{\baselineskip}
**Core Instructions:**
\par\vspace{\baselineskip}
1.  **Deconstruct the Problem:** Read the word problem carefully. Identify all the given numbers, quantities, and the relationships between them. Clearly understand what question needs to be answered.
2.  **Think Step-by-Step:** Do not try to solve the problem in a single leap. Break it down into smaller, manageable steps. Your reasoning process is more important than the final answer.
3.  **Show Your Work:** For each step, explain the logic behind it and show the calculation. For example, instead of just writing "10 - 4 = 6", write "To find the number of remaining apples, I subtract the 4 apples that were eaten from the initial 10 apples: 10 - 4 = 6."
4.  **Double-Check Your Work:** Before concluding, review your steps. Do they logically follow each other? Are the arithmetic calculations correct? Does the final answer make sense in the context of the problem?
5.  **Strict Output Format:** You MUST present your entire response in the following format. Do not add any conversational text before or after this structure.
\par\vspace{\baselineskip}
\#\#\#\# Step-by-step derivation:
1. [First logical step and calculation, explained clearly.]
2. [Second logical step and calculation, explained clearly.]
...
N. [The final step that arrives at the answer.]
\par\vspace{\baselineskip}
\#\#\#\# The final answer is: [The final numerical answer only]
} % End of \obeylines scope
\\
        \hline
    \end{tabular}
\end{table}

\clearpage
\subsection{ARC-C Prompts} \label{appendix:arccprompts}

Table \ref{tab:arcc_prompt} shows the exact system prompt used for ARC-c dataset for all the models

\begin{table}[h!] % The * tells LaTeX to span both columns
    \centering
    \caption{System prompt used for ARC-c dataset during inferencing.}
    \label{tab:arcc_prompt}
    \renewcommand{\arraystretch}{1.5}
    % Use \linewidth or \textwidth to fill the newly expanded space
    \begin{tabular}{ | p{0.95\textwidth} | }
        \hline
        \textbf{System Prompt} \\
        \hline
        % The { brackets } scope the \obeylines command just to this cell
        {\obeylines
        You are an expert AI assistant specializing in scientific reasoning and problem-solving. Your task is to meticulously analyze and answer multiple-choice questions from the AllenAI ARC dataset. These questions require deep understanding and logical reasoning, not just fact recall.

Your response must be structured in two distinct parts: your detailed **Reasoning Process** followed by the **Final Answer** in a specified format.
\par\vspace{\baselineskip}
**Part 1: Reasoning Process**
\par\vspace{\baselineskip}
Before providing the final answer, you must first work through the following structured thinking process. This section should contain your detailed analysis.
\par\vspace{\baselineskip}
1.  **Deconstruct the Question:**
    *   Identify the core scientific concept being tested.
    *   Break down the question into its fundamental components and constraints.
    *   Paraphrase the question to confirm your understanding of what is being asked.
\par\vspace{\baselineskip}
2.  **Analyze the Options:**
    *   Evaluate each multiple-choice option independently.
    *   For each option, explain the scientific principles or logic that support or refute it.
    *   Critically assess the validity of each choice in the context of the question.
\par\vspace{\baselineskip}
3.  **Synthesize and Conclude:**
    *   Provide a step-by-step chain of thought that logically connects the question's requirements to the most plausible answer.
    *   Explicitly compare the options and eliminate the incorrect ones based on your analysis, leading to your final conclusion.
    *   Always remember to encapsulate the final answer as mentioned in **Part 2**
\par\vspace{\baselineskip}
**Part 2: Final Answer**
\par\vspace{\baselineskip}
Final Answer Encapsulation: The final answer must be enclosed in \boxed{answer}.
} % End of \obeylines scope
\\
        \hline
    \end{tabular}
\end{table}

\clearpage
\section{Models Used}
\label{appendix:models}
 Detailed list of all models used with their hugging face links:

\begin{table}[htbp]
    \centering
    \caption{Overview of Parent and Child Models}
    \label{tab:model_list}
    \renewcommand{\arraystretch}{1.2}
    \begin{tabular}{@{}llcl@{}}
        \toprule
        \textbf{Category} & \textbf{Model Name} & \textbf{Size} & \textbf{HuggingFace URL} \\
        \midrule
        \textbf{Parent Models} 
        & microsoft/phi-4 & 14B & \href{https://huggingface.co/microsoft/phi-4}{Link} \\
        & Qwen/Qwen2.5-14B-Instruct & 14B & \href{https://huggingface.co/Qwen/Qwen2.5-14B-Instruct}{Link} \\
        & Mistral-7B-Instruct-v0.3 & 7B & \href{https://huggingface.co/mistralai/Mistral-7B-Instruct-v0.3}{Link} \\
        \midrule
        \textbf{Child Models} 
        & google/gemma-2-2b-it & 2B & \href{https://huggingface.co/google/gemma-2-2b-it}{Link} \\
        & google/gemma-2-9b-it & 9B & \href{https://huggingface.co/google/gemma-2-9b-it}{Link} \\
        & meta-llama/Llama-3.1-8B-Instruct & 8B & \href{https://huggingface.co/meta-llama/Llama-3.1-8B-Instruct}{Link} \\
        & meta-llama/Llama-3.2-1B-Instruct & 1B & \href{https://huggingface.co/meta-llama/Llama-3.2-1B-Instruct}{Link} \\
        & meta-llama/Llama-3.2-3B-Instruct & 3B & \href{https://huggingface.co/meta-llama/Llama-3.2-3B-Instruct}{Link} \\
        & Qwen/Qwen2-7B-Instruct & 7B & \href{https://huggingface.co/Qwen/Qwen2-7B-Instruct}{Link} \\
        & Qwen/Qwen3-0.6B & 0.6B & \href{https://huggingface.co/Qwen/Qwen3-0.6B}{Link} \\
        \bottomrule
    \end{tabular}
\end{table}

\section{Best $\lambda$ Used}
\label{appendix:best_lambda}
Below is the list of best $\lambda$ identified using validation dataset for GSM8k dataset

\begin{table}[h]
    \centering
    \caption{Model Comparison Data (Transposed)}
    \label{tab:model_data_transposed}
    \begin{tabular}{l ccc}
        \toprule
        \textbf{Child Model} & 
        \textbf{microsoft/phi-4} & 
        \textbf{Qwen/Qwen2.5-14B-Instruct} & 
        \textbf{Mistral-7B-Instruct-v0.3} \\
        \midrule
        gemma-2-2b-it & 0.6 & 0.8 & 0.04 \\
        \addlinespace
        gemma-2-9b-it & 0.01 & 0.03 & 0.9 \\
        \addlinespace
        Llama-3.1-8B-Instruct & 0.07 & 0.01 & 0.07 \\
        \addlinespace
        Llama-3.2-1B-Instruct & 0.03 & 0.08 & 0.01 \\
        \addlinespace
        Llama-3.2-3B-Instruct & 0.02 & 0.01 & 0.03\\ % Empty cell
        \addlinespace
        Qwen2-7B-Instruct & 0.5 & 0.7 & 0.01\\ % Empty cell
        \addlinespace
        Qwen3-0.6B & 0.04 & 0.04 & 0.04\\ % Empty cell
        \bottomrule
    \end{tabular}
\end{table}

\clearpage
\section{Additional Inference Time Scaling results}
\label{appendix:additional_its_results}
\begin{figure*}[htb]
\centering
% Subfigure (a)
\begin{minipage}[b]{0.45\linewidth}
    \centering
    \includegraphics[width=\linewidth]{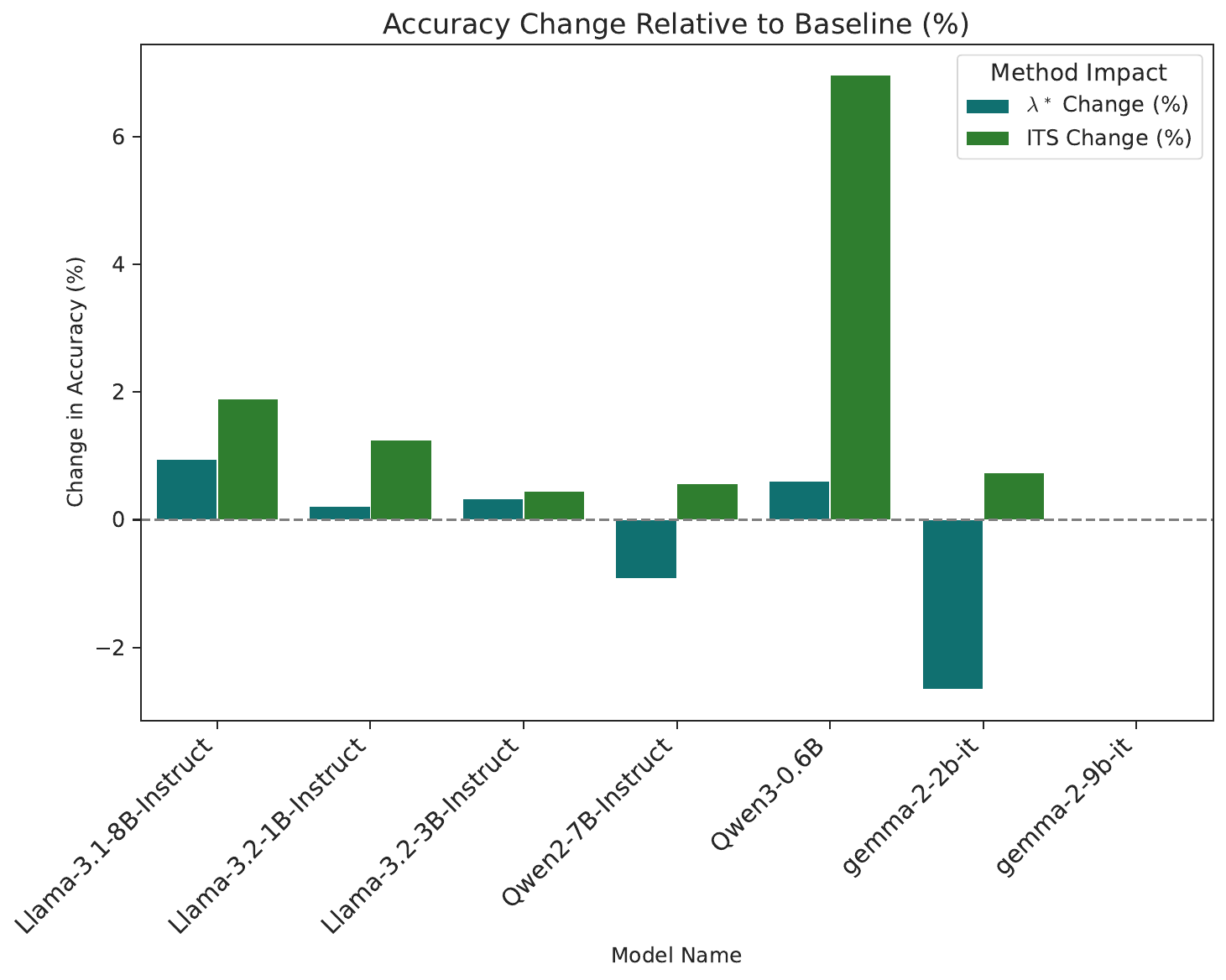}
    % \caption*{(a) Result 1}
    \vspace{5pt} % Adjust space between image and text
    \textbf{(a)}
\end{minipage}% <--- PERCENT SIGN PREVENTS NEW PARAGRAPH
\hfill
% Subfigure (b)
\begin{minipage}[b]{0.45\linewidth}
    \centering
    \includegraphics[width=\linewidth]{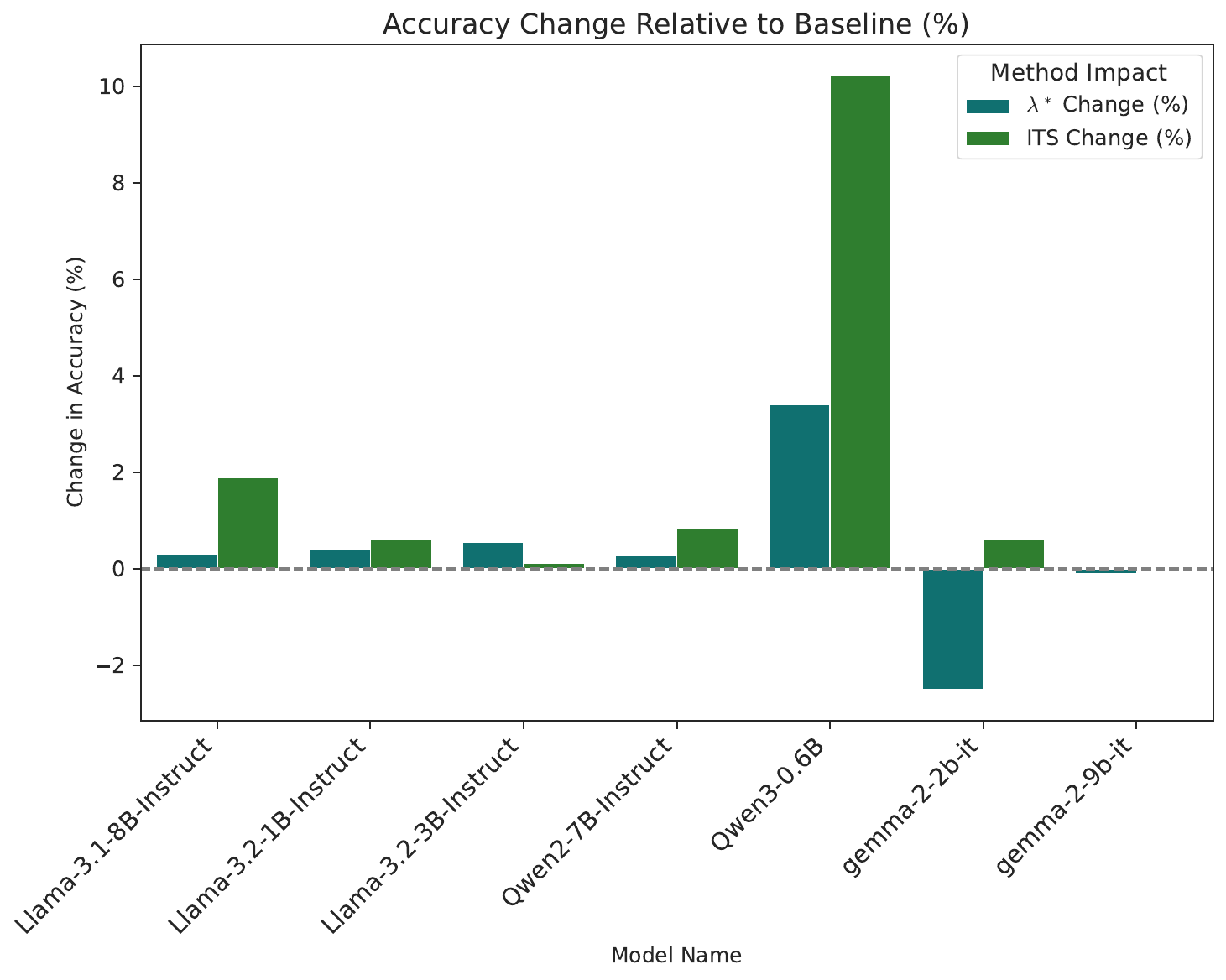}
    % \caption*{(b) Result 2}
    \vspace{5pt} % Adjust space between image and text
    \textbf{(b)}
\end{minipage}% <--- PERCENT SIGN PREVENTS NEW PARAGRAPH
\hfill
% Subfigure (c)
\begin{minipage}[b]{0.45\linewidth}
    \centering
    \includegraphics[width=\linewidth]{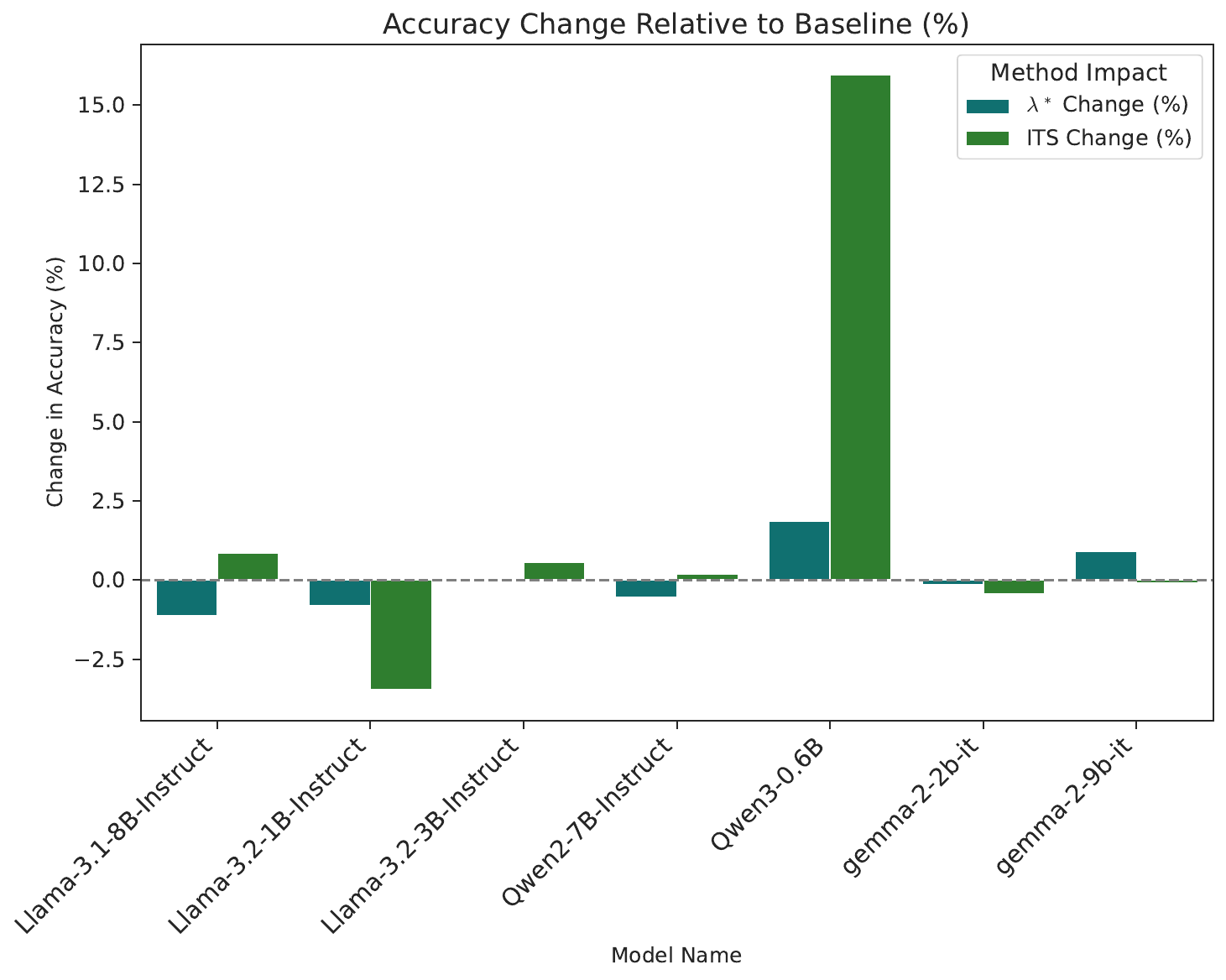}
    \vspace{5pt} % Adjust space between image and text
    \textbf{(c)}
\end{minipage}

\caption{Change in accuracy of $\lambda^*$ and Inference time scaling with baseline using multiple $\lambda$ values. (a) referes to using phi-4 as parent model, (b) refers to using Qwen2.5 14b-Instruct as parent model, (c) refers to Mistral-7B-Instruct-v0.3 as parent model}
\label{fig:acc_change}
\end{figure*}
\end{document}